\documentclass[letterpaper, 10 pt, conference]{ieeeconf} 
\IEEEoverridecommandlockouts
 
\usepackage{cite}
\usepackage{amsmath,amssymb,amsfonts}
\usepackage{algorithmic}
\usepackage{graphicx}
\usepackage{textcomp}
\usepackage{xcolor}
\usepackage{cleveref}
\usepackage{bm}
\usepackage{mathtools}
\usepackage{physics}
\usepackage{multirow}
\usepackage{siunitx}
\usepackage[nolist,nohyperlinks]{acronym}

\acrodef{RMP}{\emph{Riemannian motion policy}}
\acrodefplural{RMP}{\emph{Riemannian motion policies}}
\acrodef{MAV}{\emph{micro aerial vehicle}}
\acrodef{OMAV}{\emph{omnidirectional micro aerial vehicle}}
\acrodef{ESDF}{\emph{Euclidian signed distance field}}
\acrodef{TSDF}{\emph{truncated signed distance field}}
\acrodef{SDF}{\emph{signed distance field}}

\newcommand{\ray}{\vec{\mathbf{r}}}
\renewcommand{\vec}[1]{\mathbf{#1}}

\newcommand{\foccugrad}{\nabla\mathcal{D}}
\newcommand{\foccupdist}{\mathcal{D}}
\newcommand{\haltonseq}{\mathcal{H}}
\newcommand{\fraytrace}{\mathcal{R}}
\newcommand{\policy}{\mathcal{P}}
\newcommand{\robotposition}{\vec{x}}
\newcommand{\robotvelocity}{\vec{\dot{x}}}
\newcommand{\metric}{\mathbf{A}}
\newcommand{\polf}{\vec{f}}

\newcommand{\distray}{d}
\newcommand{\obs}{\mathbf{q}}
\newcommand{\obsvec}{\mathbf{d}_{\obs}}
\newcommand{\nicetilde}{{\raise.17ex\hbox{$\scriptstyle\sim$}}
}
\newcommand{\Rthree}{\mathbb{R}^{3}}

\def\BibTeX{{\rm B\kern-.05em{\sc i\kern-.025em b}\kern-.08em
    T\kern-.1667em\lower.7ex\hbox{E}\kern-.125emX}}
\begin{document}

\title{Obstacle avoidance using raycasting and Riemannian Motion Policies at kHz rates for MAVs}

\author{Michael Pantic{*}, Isar Meijer{*}, Rik Bähnemann, Nikhilesh Alatur, Olov Andersson, \\Cesar Cadena, Roland Siegwart, and Lionel Ott%
\thanks{* Authors contributed equally to this work.}%
\thanks{All authors are with the Autonomous Systems Lab, ETH Zürich.}
\thanks{Corresponding Author: M. Pantic, \texttt{mpantic@ethz.ch}}
 }

\maketitle
\thispagestyle{empty}
\pagestyle{empty}
\begin{abstract}
This paper presents a novel method for using Riemannian Motion Policies on volumetric maps, shown in the example of obstacle avoidance for Micro Aerial Vehicles (MAVs). Today, most robotic obstacle avoidance algorithms rely on sampling or optimization-based planners with volumetric maps. However, they are computationally expensive and often have inflexible monolithic architectures. Riemannian Motion Policies are a modular, parallelizable, and efficient navigation alternative but are challenging to use with the widely used voxel-based environment representations. 
We propose using GPU raycasting and tens of thousands of concurrent policies to provide direct obstacle avoidance using Riemannian Motion Policies in voxelized maps without needing map smoothing or pre-processing. Additionally, we present how the same method can directly plan on LiDAR scans without any intermediate map. We show how this reactive approach compares favorably to traditional planning methods and can evaluate $65536$ rays with a rate of up to $3$kHz. We demonstrate the planner successfully on a real MAV for static and dynamic obstacles. The presented planner is made available as an open-source software package\footnote{https://github.com/ethz-asl/reactive\_avoidance}.
\end{abstract}
\section{Introduction}
From the moment mobile robots could move through the world autonomously, obstacle avoidance algorithms have been fundamental.
Especially for flying robots obstacle avoidance is critical and challenging because they are usually not collision tolerant and have limited computational and sensing capabilities.
These challenges have sparked a wide variety of research in obstacle avoidance for \acp{MAV}, from sensor-based reactive controllers \cite{bouabdallah2007fullcontrol}\cite{oleynikova2015reactive} to onboard map building for sampling-based and optimization-based methods \cite{oleynikova2016continuous} to recent data-driven approaches \cite{loquercio2021highspeed}.

Early methods such as potential fields \cite{khatib1986real} are computationally efficient and versatile. However, it is not apparent \textit{how} to design such potential functions in real-world applications. Map-based approaches, e.g., occupancy maps and \acp{SDF}, filter and extract information to obtain obstacle gradients and occupancy probability. While these maps are highly effective for global  sampling-based or optimization-based planning, the typical mapping-planning cycle is computationally expensive, relies on accurate state estimates, and introduces aliasing and delays. Additionally, there is the risk that the map is simply wrong or outdated, giving rise to the need for a navigation paradigm that can effortlessly combine map and live sensor data.
While end-to-end methods have shown great promise and robustness recently, they often suffer in generalizability, introspectability, and modularity.

A promising class of navigation algorithms is \acp{RMP}~\cite{ratliff2018riemannian}.
Instead of formulating a monolithic, single navigation algorithm, the theory behind \acp{RMP} proposes breaking up the navigation problem into many individual policies. These policies are typically relatively simple, as they describe an action in the geometric space where the problem is the easiest to solve. However, combining many seemingly simple policies can exhibit complex overall robot behavior, while providing a high degree of modularity.
Most works using \acp{RMP} focus on settings centered around (self-intersections of) robotic arms or structured environments, where idealized environment representations, such as meshes or primitives, are available.

So far, planning research has not shown to use \acp{RMP} efficiently on voxelized maps in a realistic onboard setting. Earlier work considers a single policy to the closest obstacle \cite{meng2019neural} or needs to convert the voxelized map to a different representation, such as a geodesic field \cite{mattamala2022reactive}.
%
\begin{figure}[bt]
\centering
  \includegraphics[width=0.48\textwidth]{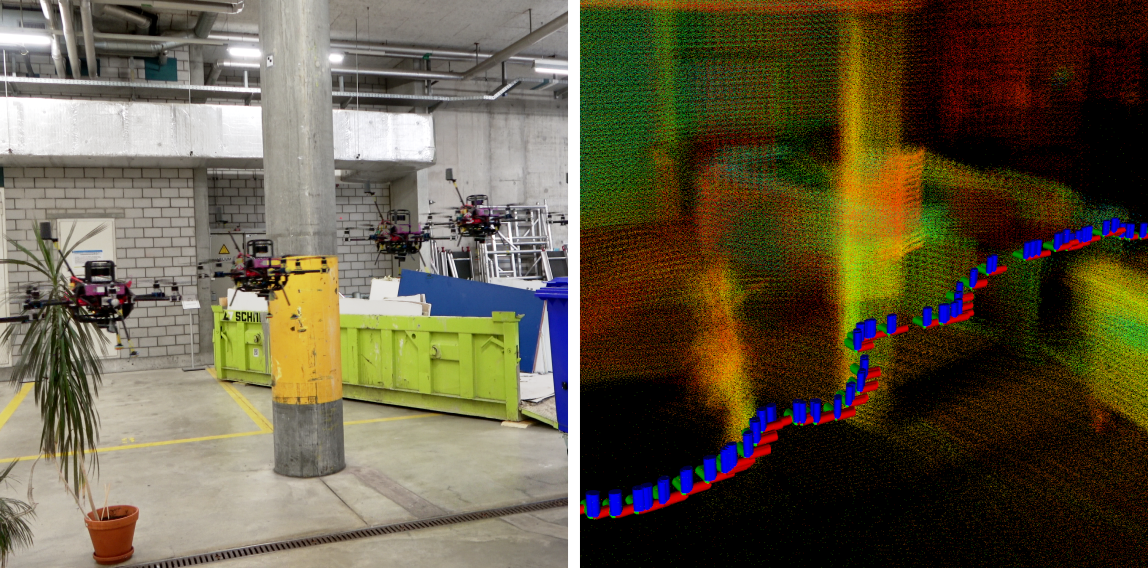}
  \caption{Left: An \ac{MAV} avoiding obstacles using the proposed navigation algorithm without a map, but raw LiDAR data instead. We only command the desired goal location. Right: Accumulated point cloud for visualization and the executed trajectory.}
  \label{fig:fig_garage}
  \vspace{-1.5em}
\end{figure}
This paper presents and investigates highly parallelizable Riemannian Motion Policies for obstacle avoidance in unstructured environments on \acp{MAV} on voxel-based map representation. To this end, we contribute:
\begin{itemize}
    \item a novel raycasting-based policy generation method capable of evaluating thousands of policies in parallel on 3D voxel maps;
    \item an extension to synthesize motion policies directly from LiDAR data;
\end{itemize}
Additionally, the proposed navigation algorithms are evaluated on actual flight experiments and made available open-source. The modular nature of \acp{RMP} facilitates the easy integration of new policies and adjusting to novel scenarios, with the proposed policies providing building blocks for further research and practical applications.
\section{Related work}
In the following, we present relevant related work focusing on applying \acp{RMP} for MAV navigation in unstructured environments.
Early approaches for reactive obstacle avoidance on MAVs combined simple ultra-sound or infrared distance sensing with attitude control \cite{bouabdallah2007fullcontrol} or handcrafted algorithms \cite{Gageik2015obstacle} to steer the MAV away from obstacles. While these early approaches are robust, reactive, and power efficient, they do not scale well regarding the number of measurements. Later, online stereo depth estimation algorithms made reactive collision avoidance at higher resolution possible \cite{oleynikova2015reactive}. Existing reactive, sensor-based approaches often operate at the controller level. Data-driven obstacle avoidance methods recently have shown great robustness and versatility, especially for fast and dynamic flight \cite{loquercio2021highspeed}. However, all previously presented methods are monolithic designs requiring complete re-design or re-training for additional use cases or new requirements.

The second significant part of research on collision avoidance for MAVs revolves around online mapping and planning using a volumetric map representation. Octomap \cite{hornung2013octomap} and voxblox \cite{oleynikova2017voxblox} are standard volumetric occupancy mapping systems used for planning. Voxblox additionally provides \ac{SDF} information and obstacle gradients, which can improve planning speeds and quality at the expense of additional processing during mapping. Sampling-based algorithms such as RRT*\cite{karaman2011sampling}, derived methods, and optimization-based methods such as CHOMP \cite{zucker2013chomp} will find efficient and safe paths in voxel-based maps. Combined with a depth sensor and an odometry/slam framework, a closed-loop obstacle avoidance system with global and local planning is possible~\cite{oleynikova2016continuous,mohta2018fastflight}. The resulting closed-loop planning rates are only a few hertz. The robustness of the solution hinges on the map quality, which is limited by state estimation drift, voxel resolution, and compute resources for sampling, collision checking, and optimization.

Riemannian Motion Policies \cite{ratliff2018riemannian} is a modular and fast framework for navigation and motion synthesis. Second-order differential equations encode behaviors on task-specific manifolds that are combined efficiently to generate the overall robot motion.
The original work \cite{ratliff2018riemannian} uses collision meshes with applications for robotic manipulators. \acp{RMP} have use cases such as navigating cars based on sensor and learned policies \cite{meng2019neural}, steering legged robots via extracting geodesic fields from height maps \cite{mattamala2022reactive} and navigating MAVs with meshes\cite{pantic2021mesh} or NeRFs\cite{pantic2022sampling} as world representation.
To the best of our knowledge, no method currently exists that allows the usage of voxel-based map representations directly with \acp{RMP}.
\section{Method}
The challenge in using voxel-based map representations with \acp{RMP} is that, unlike with meshes or geometric primitives, one can not derive a single efficient policy that induces a collision-avoiding behavior. To overcome this, we propose leveraging modern GPUs' parallelism to develop a massively parallel policy evaluation system. We combine raycasting in a voxel-based map with per ray obstacle avoidance policies directly on the GPU, enabling the efficient evaluation of up to 400 million individual ray policies per second.
Compared to approaches using single location or \ac{SDF} lookups, this use of many policies enables the navigation algorithm to factor in more varied information about the local geometry.

The following shows how such massive parallelism can be obtained and used for navigation with Riemannian Motion Policies.
Before describing our method, we first provide a brief introduction to \acp{RMP}. For more details, please refer to the original work \cite{ratliff2018riemannian}.
\subsection{Riemannian Motion Policies}
One of the main ideas behind Riemannian Motion Policies is to formulate a potentially complex behavior as a combination of many small and straightforward behaviors. The designer has the freedom to formulate these policies in whichever space they are easiest to describe. 
The essential building block is a policy $\policy$ consisting of a function $f(\robotposition,\robotvelocity)$ and accompanying Riemannian metric $\metric(\robotposition,\robotvelocity)$, where $\robotposition \in \Rthree$ is the robot's current position in the policy's coordinate frame. The function $\polf$ defines the instantaneous acceleration while the Riemannian metric $\metric$ corresponds to the weight of the policy, which can be isotropic (identical in all axes) or directional.
A single policy $\policy_C$ combines multiple policies $\lbrace\policy_0, \dots, \policy_N\rbrace$,
\begin{equation}
\mathcal{P}_c = \left(\polf_{c}, \metric_{c}\right) = \left( \left( \sum_i \metric_i \right) ^+ \sum_i \metric_i \polf_i ,\, \sum_i \metric_i \right).
\label{eq:combine_rmps}    
\end{equation}
\Cref{eq:combine_rmps} optimizes the geometric-weighted mean of all summed individual policies.
Note that there is no explicit optimizer but rather an implicit gradient-descent-like optimization by the physical system evaluating and executing the policy at every time step.
The policy combination enables fast and efficient navigation algorithms. However, as there is no time correlation and the system is purely reactive, the policies need to provide stability and smoothness through adequate design.
\subsection{Problem description and notation}
\newcommand{\robotacc}{\ddot{\robotposition}_k^{cmd}}
Our goal is to define an acceleration policy
\begin{equation}
    \policy^k\,=\,(\polf^k(\robotposition^k, \robotvelocity^k), \metric^k(\robotposition^k, \robotvelocity^k))
\end{equation}
that steers a robot from a start to a goal state while staying clear of obstacles.
At every time step $k$, our method evaluates the policy as a function of the robot's state consisting of its position $\robotposition^k$ and velocity $\robotvelocity^k$. It then applies the resulting acceleration $\ddot{\robotposition}^k =  \polf(\robotposition^k, \robotvelocity^k)$ to the robot.
Without loss of generality, we assume a 3D world and 3D policies in the following, i.e., $\robotposition, \robotvelocity, \polf \in \Rthree$ and $\metric \in \mathbb{R}^{3 \times 3}$.
Note that both $\polf^k$ and $\metric^k$ can absorb arbitrary information about the current map, sensor data, or environment -- which we here express by the explicit time dependency of both functions but omit in the following for brevity.
%
%

\subsection{Base policies}
\label{sec:base_policies}
We use the attractor and obstacle avoidance policies defined in \cite{ratliff2018riemannian} as basic building blocks.
An attractor policy creates an attractive force and can be used to define goal locations. Let $\robotposition_a$ denote the target towards which the robot should move. The attractor policy $\policy_{a} = (\polf_a, \metric_a)$ is then defined as 
\begin{equation}
\label{eq:attractorPolicy}
\begin{aligned}
    \polf_a(\robotposition, \robotvelocity) &= \alpha_a \bm{s} \left( \robotposition_a - \robotposition \right) - \beta_a \robotvelocity \\
    \metric_a(\robotposition, \robotvelocity) &= \mathbb{I}^{3 \times 3}
    \end{aligned}\;,
\end{equation}
where $\alpha_a, \beta_a > 0$ are gain respectively damping scalar parameters. We define the soft normalization function $\bm{s}$ as:
\begin{equation}
\label{eq:softmax}
\bm{s}\left(\bm{v}\right) = \frac{\bm{v}}{\norm{\bm{v}} + c \log \left(1 + \exp \left(-2c\bm{v}\right) \right)},\,\, c > 0\;,
\end{equation}
where $c$ is a tuning parameter.
The second building block, the obstacle avoidance policy $\policy_{obs} = (\polf_{obs}, \metric_{obs})$, is the summation of a repulsive term $\polf_{rep}$ and a damping term $\polf_{damp}$:
\begin{equation}
    \polf_{obs} \left(\robotposition, \, \robotvelocity, \, \ray,\, \distray \right) = \polf_{rep}\left(\robotposition,\,\ray,\,\distray\right) + \polf_{damp}\left(\robotvelocity,\,\ray,\,\distray\right)\;,
\end{equation}
where $\ray$ is the unit vector pointing away from the obstacle, and $\distray$ is the distance to the obstacle, both functions of the robot's position. The repulsive term is:
\begin{equation}
    \polf_{rep}\left(\robotposition,\,\ray,\,\distray\right) = \eta_{rep}\exp \left(- \frac{d}{\upsilon_{rep}}\right) \ray \:,
\end{equation}
where $\eta_{rep}, \, \upsilon_{rep} > 0$ are a gain and a length scaling factor, respectively. Next, we define the vector $\mathbf{g}_{obs}$ that points away from the obstacle, scaled according to the velocity projected onto $-\obsvec$, and vanishing to 0 if the velocity vector points away from the obstacle.
\begin{equation}
    \mathbf{g}_{obs}\left(\robotvelocity,\, \ray\right) = \max \left\{0, -\robotvelocity^T\ray\right\}^2 \ray \:,
    \label{eq:obs_function}
\end{equation}
 Using \eqref{eq:obs_function} the damping term is:
\begin{equation}
    \polf_{damp}\left(\robotvelocity,\,\ray,\,\distray\right) = \eta_{damp} \bigg/ \left(\frac{\distray}{\upsilon_{damp}} + \epsilon\right) \cdot \mathbf{g}_{obs}\left(\robotvelocity, \, \ray\right) \:,
\end{equation}
where $\eta_{damp}$ and $\upsilon_{damp}$ are gain and length scaling factors, and $0 < \epsilon \ll 1$ ensures numerical stability. The resulting metric $\metric_{obs}$ is:
\begin{multline}
\label{eq:metric}
    \metric_{obs}\left(\robotposition, \, \robotvelocity, \, \ray,\, \distray\right) = \\  w_r\left(\distray\right) \cdot \bm{s}\left(\polf_{damp}\left(\robotvelocity,\, \ray,\, \distray\right)\right)
    \bm{s}\left(\polf_{damp}\left(\robotvelocity,\, \ray,\, \distray\right)\right) ^T,    
\end{multline}
where the soft-normalization function $\bm{s}\left(\cdot\right)$ is the same as \eqref{eq:softmax}, and $w_r\left(\cdot\right)$ is a weight function that defines the radius $r$ in which the policy is active. This radius is defined as $w_r\left(d\right) = \frac{1}{r^2}d^2 - \frac{2}{r} d + 1$, which is a cubic spline that goes from $w_r\left(0\right) = 1$ to $w_r\left(r\right) = 0$, with derivatives equal to 0 at the endpoints. For distances greater than $r$, the weighing function returns 0. This metric is directionally stretched towards the obstacle, indicating that the policy acts stronger in this direction.
\subsection{Simple obstacle avoidance in ESDF maps}
\label{chap:esdf_policy}
Voxel mapping frameworks such as voxblox \cite{oleynikova2017voxblox} that carry an \ac{ESDF} representation allow querying the distance to the closest obstacle and the corresponding obstacle gradient.
We define the function that returns the distance to the closest obstacle given a position as 
$\foccupdist(\robotposition) \in \mathbb{R}^{+}$, with the corresponding obstacle gradient $\foccugrad(\robotposition) \in \Rthree$.
A simple single-query obstacle avoidance policy would use $\foccupdist(\robotposition)$ as $\distray$ and $\foccugrad(\robotposition)$ as $\ray$ for the obstacle avoidance policy defined in Section \ref{sec:base_policies}.
Together with a goal attractor policy, this constitutes a simple local planner. Such a simple approach, however, in practice suffers from poor stability. To address this, the next section introduces our proposed raycasting method.
\subsection{Raycasting based obstacle avoidance}
\label{chap:ray_policy}
\begin{figure}[bt]
\centering
  \includegraphics[width=\linewidth]{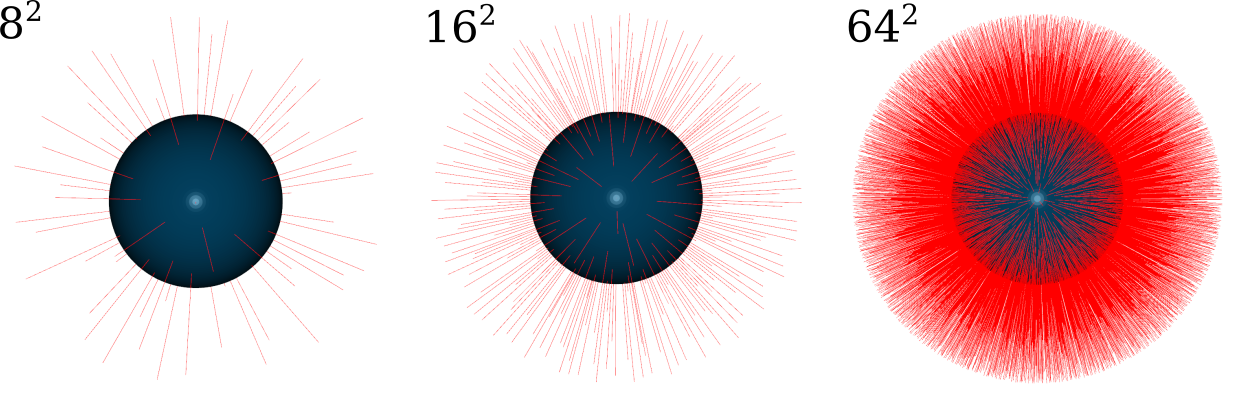}
  \caption{Visualization of the Halton sequence sampling mapped to spherical coordinates for different numbers of samples ($8^2, 16^2, 64^2)$.}
  \label{fig:halton_seq}
\end{figure}

In order to efficiently use as much spatial information as possible at every time step, we use many parallel policies combined with efficient map queries through raycasting.
The core idea is to generate a set of rays $\left\{\ray_0, \dots, \ray_{N}\right\}$, along which we perform raycasting to determine the distance to an obstacle.
To generate a sampling of $N+1$ quasi-random, equally spaced rays radiating from the robot's center, we use a 2D Halton sequence $\haltonseq(\cdot)$ \cite{wong1997sampling} with base $2$ for sampling the azimuth $\varphi_{i}$, and base $3$ for elevation $\theta_{i}$:
\begin{equation}
    \begin{aligned}
       \varphi_i & = \cos^{-1}(1-2\,\haltonseq(i,2)) \\
       \theta_i  & = 2\,\pi\,\haltonseq(i,3)
    \end{aligned}
    \qquad \forall i = 0,\dots,N
    \label{eq:halton}
\end{equation}
The resulting ray is then transformed into a unit direction vector
$\ray_{i} = (
        \sin(\varphi_{i}) \cos(\theta_{i}),
        \sin(\varphi_{i}) \sin(\theta_{i}),
        \cos(\varphi_{i})
    )$.
\Cref{fig:halton_seq} shows an increasing number of rays sampled in this manner.
For each ray $\ray_i$, we calculate the distance to the nearest obstacle from the current robot position $\robotposition$ along the ray using the raycasting function $\distray_{i} = \fraytrace(\robotposition, \ray_{i}) \in \mathbb{R}^{+}$.
The complete raycasting-based obstacle avoidance policy $\policy_{ray}$ according to \cref{eq:combine_rmps} is 
\begin{equation}
\begin{aligned}
    \metric_{ray}(\robotposition, \robotvelocity) &= \sum_{i=0}^{N}\, \metric_{obs}(\robotposition, \robotvelocity, \ray_{i},\distray_{i}) =  \sum_{i=0}^{N} A_{obs}^{i}  \\
    \polf_{ray}(\robotposition, \robotvelocity) &= \metric_{ray}^{+}  \, \sum_{i=0}^{N} \metric_{obs}^{i} \Bigl( \polf_{rep}(\robotposition, \ray_{i}, \distray_{i})\,+\\
    &\quad\quad\quad\quad\quad\quad \polf_{damp}(\robotvelocity, \ray_{i},\distray_{i})\Bigr).
\end{aligned}
\label{eq:raytracing_policy}
\end{equation}
The complete raycasting policy $\policy_{ray}$ provides obstacle avoidance that is direction, distance, and speed dependent. It has a 0 metric for all rays pointing sufficiently away from the robot's velocity, effectively also disabling the repulsive component for these rays. All close rays that are in the direction of the robots velocity, i.e., posing a risk for collision, generate a large metric and acceleration and therefore drive the robot away.

The summation of the raycasting policy $\policy_{ray}$ with a suitable goal attractor policy $\policy_{a}$  produces a local planning policy that drives the robot to the desired goal and keeps it clear of obstacles.
 The acceleration command sent to the robot is the geometric weighted mean (\cref{eq:combine_rmps}) of both policies. Movement and reiteration implicitly optimizes the joint objective of both policies.
While this approach is conceptually simple, it requires the evaluation of thousands of raycasting policies, which is computationally expensive. However, as described in the next section, the inherently parallel nature of \acp{RMP} combined with a modern GPU implementation of a voxel-based map \cite{nvblox} allows us to offload the planning process to the GPU.
\subsection{Implementation}
An efficient raycasting policy implementation is crucial to reach the required refresh rates for fast and reactive navigation.
\begin{figure}[bt]
    \centering
  \includegraphics[width=0.4\textwidth]{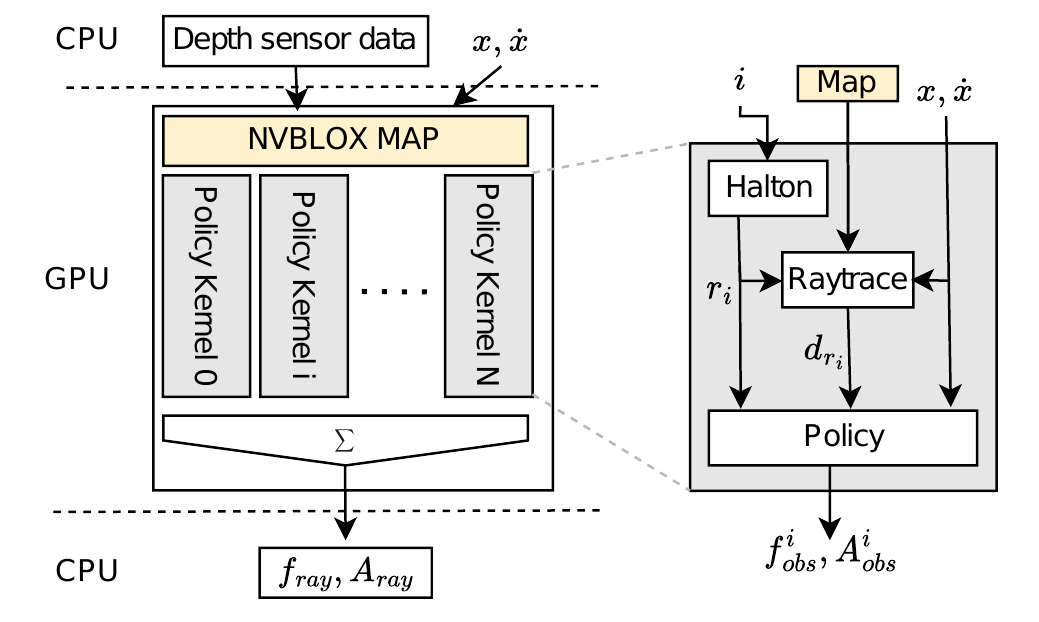}
  \caption{Left: Overview of the overall execution architecture. The map is kept on the GPU, for each ray an individual policy kernel is started. The summation $\Sigma$ (\cref{eq:combine_rmps}) is implemented using an efficient block-reduce paradigm. Right: Data flow diagram of an individual policy kernel. As the policy kernels have no interdependency they parallelize efficiently.}
  \label{fig:halton}
  \vspace{-1.5em}
\end{figure}
We implement a single CUDA kernel that combines ray computation based on thread id and the Halton sequence, raycasting, and policy calculation.
The raycasting is adopted from nvblox \cite{nvblox} and uses the \ac{SDF} and voxel information to step through free space in a loop efficiently.
Each ray policy executes as an instance of this kernel on a CUDA thread.
The map of the environment, implemented using nvblox \cite{nvblox}, is kept in GPU memory and shared between all kernels, meaning that the kernels do not require additional GPU transfers during execution. 

The ray and policy operations do not have any cross kernel inter-dependencies except the access to global shared GPU memory, leading to almost linear scaling w.r.t. the number of computing cores available.
A block-reduce scheme combines all the individual policies into a single policy as per equation \cref{eq:combine_rmps}. Finally, we only transfer the resulting policy and its metric, $12$ float values, respectively $48$ bytes, back to the host computer for execution.
\subsection{LiDAR scan policies}
The raycasting paradigm for obstacle avoidance is a natural fit for raw LiDAR data, where each LiDAR beam represents the result of a physical raycasting operation.
Instead of using the Halton sequence $\haltonseq$ and raycasting function $\fraytrace$ inside a map, we directly use the LiDAR beam direction and sensed range for every LiDAR beam in a laser scan.
Due to the highly efficient parallelization, we can execute a sufficient amount of policies to address every LiDAR beam in every scan produced by the LiDAR. 
For example, evaluating an Ouster OS-0 128 beam LiDAR scan with 2048 rotational increments corresponds to evaluating \num{262144} individual obstacle avoidance policies every \SI{0.1}{\sec}, which processes all available information about the local geometry.
\section{Experimental setup}
\label{sec:experiments}

In the following, we evaluate our proposed system quantitatively in simulation to demonstrate the ability to navigate successfully in complex 3D environments. We vary the number of local policies used at each time step and compare the results to an \ac{ESDF}-based policy described in \Cref{sec:base_policies}, similar to~\cite{meng2019neural} and to CHOMP~\cite{zucker2013chomp}. Next, we provide run-time information on different hardware configurations. Finally, we present qualitative results of the method running on a real platform. The parameters used throughout the experiments are listed in \Cref{tab:parameters}.
\subsection{Map-based evaluations}
\begin{figure}[bt]
\centering
  \includegraphics[width=0.5\textwidth]{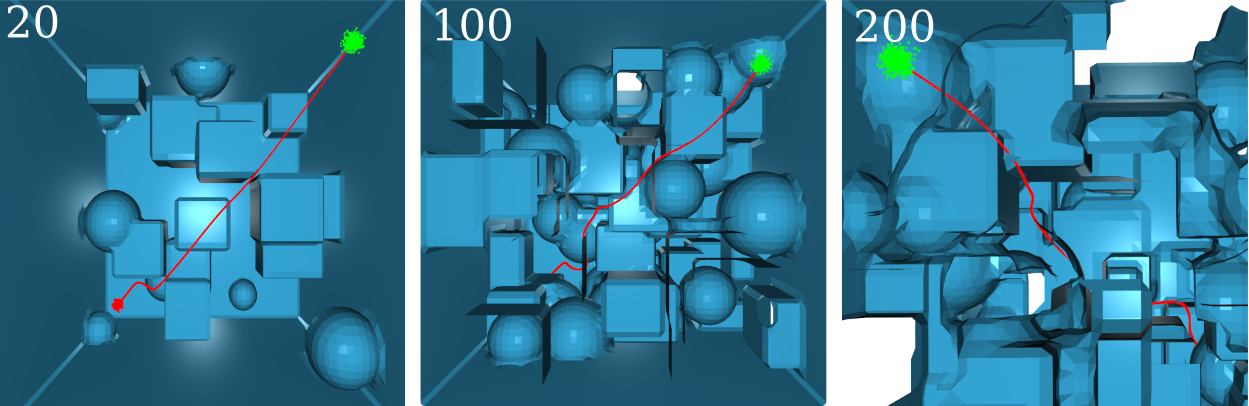}
  \caption{Rendering of exemplary random maps with different number of obstacles within the same $10 \times 10 \times 10\, [m]$ perimeter. The green and red points indicate desired start and end-points for trajectories. The red trajectories are generated using the RMP raycasting policy with $64^{2}$ rays. Note that the $200$ map appears less dense than it actually is, as not all obstacles are shown for visualization (backface culling).}
  \label{fig:maps}
\end{figure}

To provide a balanced and statistically meaningful evaluation of the raycasting policy described in \cref{eq:raytracing_policy}, we use a large and diverse amount of randomly generated maps.
To generate these maps, we uniformly sample geometric primitives such as spheres and boxes and merge them through boolean operations. \Cref{fig:maps} shows three instances of dense, cluttered maps with increasing primitive numbers. The resulting geometry has high variability, as intersections and fusions generate small and intricate but also large features. We use a maximum of $200$ obstacles corresponding to about $45\%$ occupied voxels on average.

The main tuning parameter of the raycasting policy that greatly influences the execution speed and query resolution is the number of rays sampled. How many should one use in general, and where do diminishing returns begin? To gain insight into this question, we vary this parameter between $4^2$ = $16$ and $256^2$ = \num{65536}. Another aspect is the comparison against the naive but obvious baseline \ac{ESDF}-based policy described in \cref{chap:esdf_policy}, which only needs a single geometrical lookup per time step.
Another question we aim to answer is how well a purely reactive method can perform against an optimization-based method in terms of planning quality, time, and success rate for obstacle avoidance.
Therefore, we compare against different tunings of CHOMP \cite{zucker2013chomp}, a trajectory optimization method.
\subsection{LiDAR policy}
Due to its reactive and sensor-based nature, we present the closed-loop LiDAR variant of the proposed navigation method on an actual \ac{MAV}.
The \ac{MAV} carries an NVidia Jetson Xavier NX and an Ouster OS-1 64 beam LiDAR outputting point clouds of size $512 \times 64$ at \SI{20}{\hertz}. All processing is performed onboard and is purely sensor-based. The navigation system consists of just the LiDAR and goal policy executed at a control frequency of \SI{100}{\hertz}. The resulting acceleration is the next setpoint. The reactive nature of this setup allows for avoiding moving obstacles.
We perform two experiments with static obstacles in different environments; a large garage, and an office-sized room. We command the \ac{MAV} to reach a position behind the obstacles via the goal attractor policy.
\begin{figure}[bt]
\centering
  \includegraphics[width=0.5\textwidth]{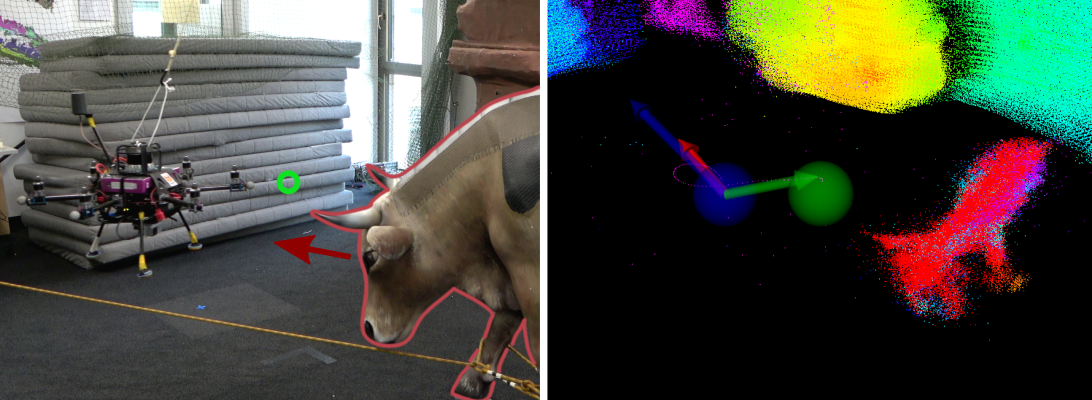}
  \caption{Left: Moving obstacle experiment. The \ac{MAV} is commanded to hold the position indicated by the green circle, while an obstacle is pulled through the environment. Right: Visualization of the policy in the same experiment, accumulated over 10 frames for visualization. The color signifies the influence of each point, where red is strongest, purple weakest. The policy's directional metric then minimizes the influence of points facing away from the \ac{MAV}s motion. The green sphere is the commanded and the blue sphere the current location of the \ac{MAV}. The green arrow is the instantaneous acceleration of the goal policy, the blue arrow the combined effect of the ray policy and the red arrow the final combination executed by the \ac{MAV}, after taking into account the directional metrics. }
  \label{fig:moving_exp}
  \vspace{-1.5em}
\end{figure}
Additionally, we run an experiment with a large moving obstacle where we commanded the \ac{MAV} to stay at a specific location in the path of the moving obstacle.
%
\begin{table}[bt]
    \centering
    \begin{tabular}{r|l|r|r}
         \textbf{Param} & \textbf{Description} & \textbf{Static Map} & \textbf{LiDAR} \\
         \hline
       $\alpha_g$ & Attractor gain & $10$ & $0.8$ \\
       $\beta_g$ & Attractor damping & $15$& $1.6$ \\
       $c$ & Softmax parameter & $0.2$& $1.0$ \\ 
       $\eta_{rep}$ & Obstacle repulsive gain & $88$ & $1.2$\\ 
       $\upsilon_{rep}$ & Obstacle repulsive scaling & $1.4$ & $1.5$ \\
       $\eta_{damp}$ & Obstacle damping gain & $140$ & $3.0$ \\ 
       $\upsilon_{damp}$ & Obstacle damping scaling & $1.2$ & $1.0$\\
       $r$ & Policy radius & $2.4$ & $1.3$ \\
        \end{tabular}
    \caption{Tuning of the obstacle avoidance policy for static map and LiDAR evaluations.}
    \label{tab:parameters}
\end{table}
\section{Results}
In the following, we present the results for the map-based statistical experiments and the sensor-based \ac{MAV} experiments.
\subsection{Map-based raycasting policy}
\begin{figure}[bt]
\centering
  \includegraphics[width=0.45\textwidth]{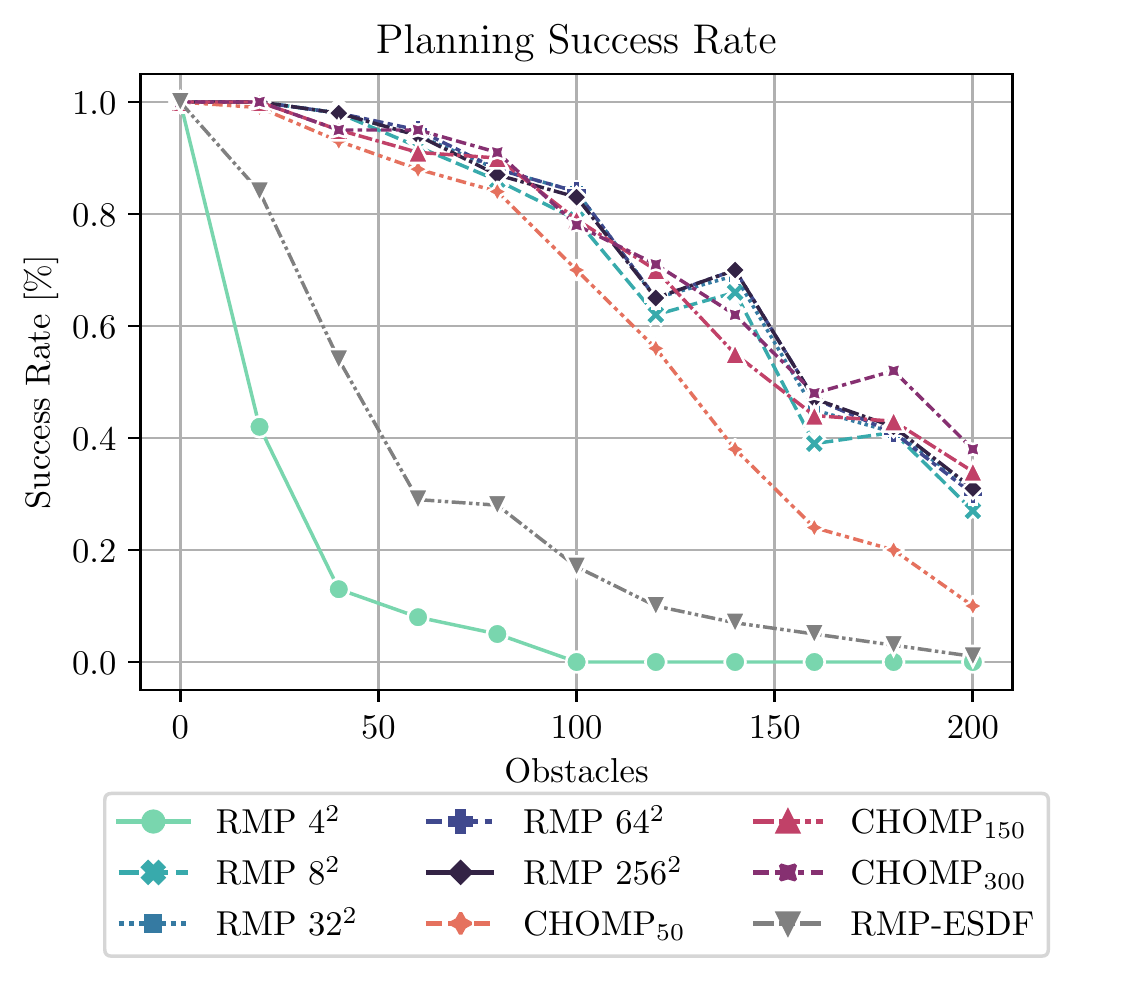}
  \caption{Comparison of success rates of different local planners. RMP $N^2$ denotes the proposed planner with $N^2$ rays generated by Halton sampling, raytraced and evaluated in the policy. For CHOMP variants the number indicates the maximum number of iterations allowed.}
  \label{fig:ray_cast_performance}
  \vspace{-1.5em}
\end{figure}

\Cref{fig:ray_cast_performance} shows the success rates of the different planners in increasingly cluttered maps. We calculate the success rate by running $100$ randomized repetitions for each planner-map combination, resulting in $\nicetilde9k$ total runs. We label a planning instance successful if it generates a collision-free path that connects the start and goal coordinates.
The proposed raycasting approach performs similarly to a well-tuned CHOMP implementation, while the naive \ac{ESDF} policy often gets stuck in local minima or collides with geometry. We attribute this to the missing damping term -- the policy can not react to more distant geometry as it only perceives the local obstacle gradient and distance.

The number of rays cast also influences on the success rate. A sampling of $4^{2}=16$ rays is too sparse to obtain good coverage of all directions. At around $32^{2}=1024$ rays, the success rate approaches the maximum observed. However, collision avoidance in filigree environments or robots with intricate geometry will benefit from even denser raycasting.
\Cref{fig:maps} shows successful paths obtained with the raycasting policy. 
While a single raycasting policy is comparably simple, a complex behavior emerges from combining many simple policies.
Even in dense and complicated maps, the generated trajectories are smooth and free of collisions.

\begin{figure}[bt]
\centering
  \includegraphics[width=0.4\textwidth]{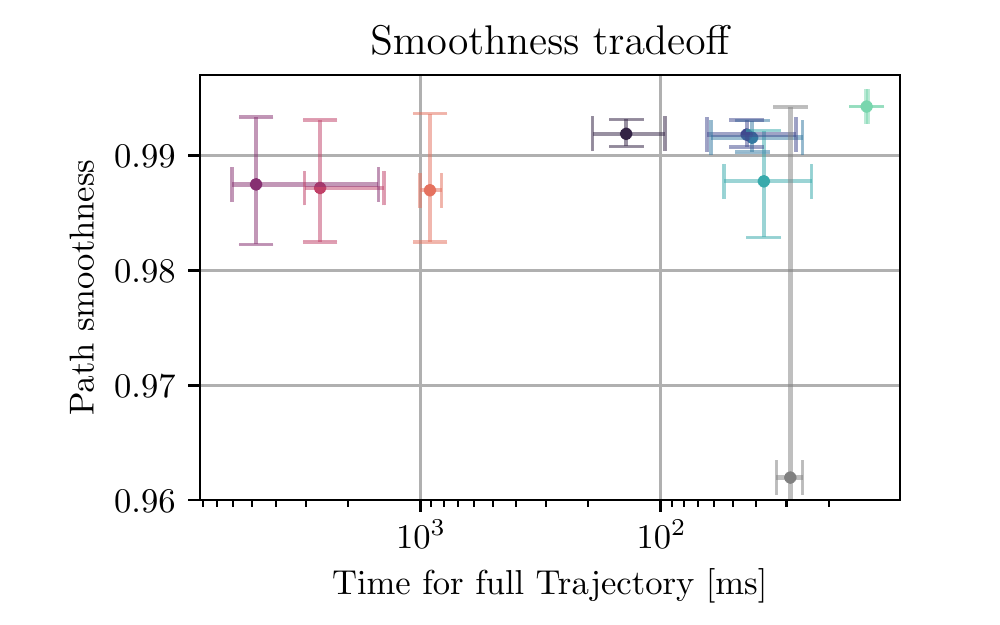}
 
  \caption{Visualization of the trade-off between trajectory smoothness and time spent for planning. Statistics from \num{5159} successful runs, with the colouring as in
  \cref{fig:ray_cast_performance}. The top right corner is the optimum, and the error-bars visualize the $10^{th}$ and $90^{th}$ percentile. Path smoothness is the averaged angular similarity between subsequent trajectory segments ($1-\frac{1}{\pi}\ cos^{-1}\left[A \cdot B/(|A||B|)\right]$), where 1.0 signifies no angular change and 0.5 signfies an average $90^{\circ}$ turn after each segment. Values below $0.95$ are visible jagged.}
  \label{fig:planner_smoothness}
    \vspace{-1.5em}
\end{figure}

\Cref{fig:planner_smoothness} visualizes the trade-off between path smoothness and the time needed to plan a complete trajectory. We observe that the proposed raycasting approach results in smooth trajectories requiring significantly less time than CHOMP. CHOMP uses an explicit smoothness term in its optimization of the spatiotemporal trajectory. For \acp{RMP}, double acceleration integration yields the trajectory.
Here, the acceleration is the average of a function of many rays, which implicitly creates a smooth progression of accelerations. Our results show smoothness greater than $0.98$ with $8^2$ rays and more. Note that the smoothness result for $4^2$ is irrelevant because this variant was mostly successful in simple environments. Here, the solution is straight trajectories which is inherently smooth.
The \ac{ESDF} policy's resulting acceleration depends only on two properties read from the map at each time step, causing higher variations in acceleration and less smooth paths.
\subsection{Reactive sensor-based navigation}
The system performs obstacle avoidance onboard at sensor frequency as expected. The policy steers the \ac{MAV} smoothly despite imperfections in the raw input point clouds, such as outliers, range noise, and occlusions. We do not perform any point cloud pre-processing or filtering except excluding points on the vehicle with a small range and intensity.
\Cref{fig:fig_garage} shows an example of the trajectory executed by the \ac{MAV} in the large garage scenario. As is visible, the system could avoid structured (pillar) and unstructured obstacles (plant). 

Moving obstacle avoidance is shown in \cref{fig:moving_exp}. The planner is robust against partial occlusion and partially incomplete point clouds due to absorption. Local minima occasionally cause the planner to get stuck - which is expected for a \textit{local} collision avoidance system. Thanks to the modular architecture, adding additional policies for \textit{global} planning is straightforward.
The navigation stack used on the order of $2\%$ of the GPU resources (at maximal $\SI{1.1}{\giga\hertz}$ GPU frequency) and $20\%$ of one ARM CPU core of the Jetson NX. The CPU load is mainly attributed to point cloud I/O.
\subsection{Benchmarks}
We benchmarked all policies on three different GPU variants. An NVidia Jetson NX with 384 CUDA cores, an NVidia MX150 with 384 cores, and a Titan XP with 3840 cores. The Jetson represents what an \ac{MAV} can carry in terms of GPU computing, while the MX150 is a standard notebook GPU.
\begin{figure}[ht]
    \centering
  \includegraphics[width=0.4\textwidth]{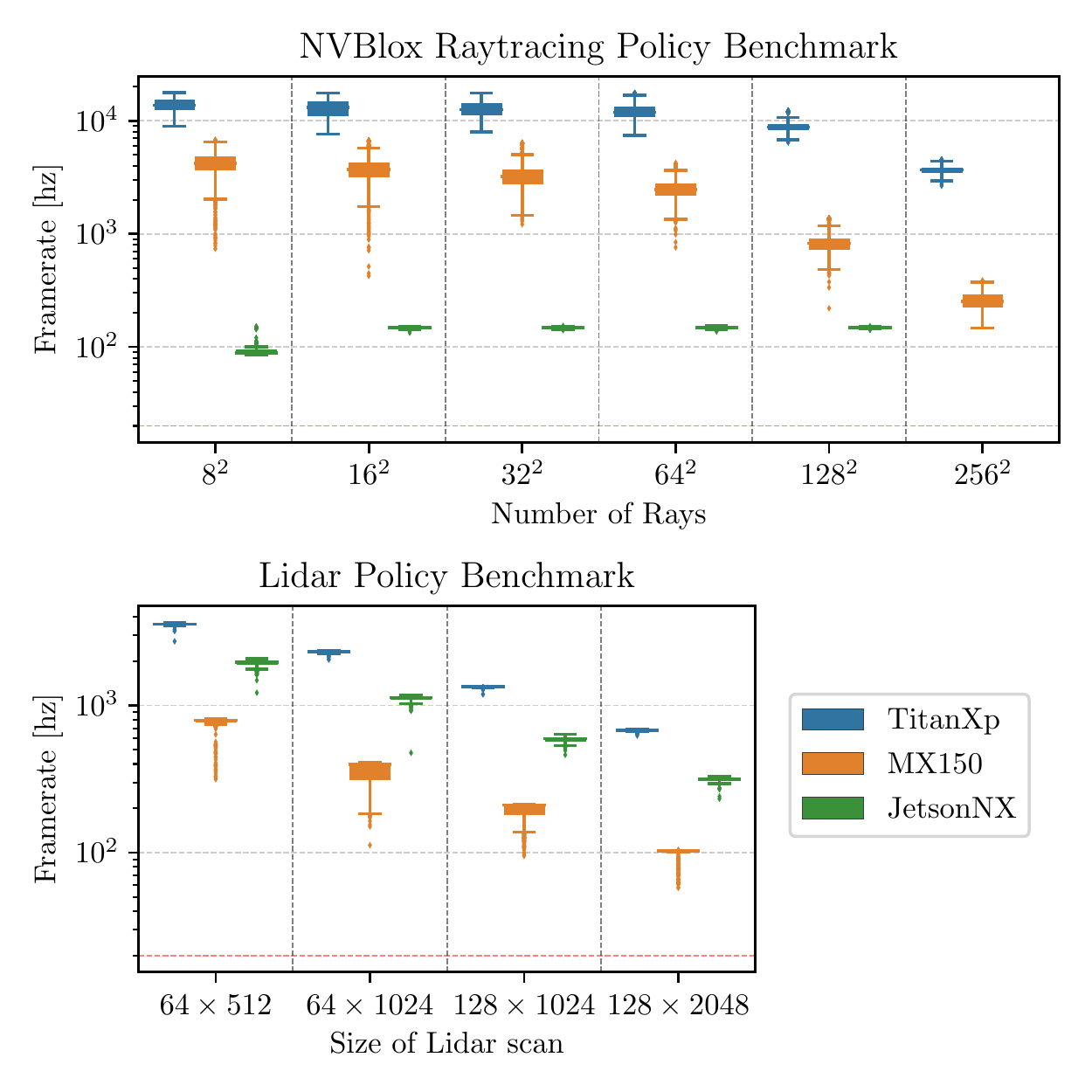}
  \caption{Benchmark statistics for policy evaluation. The framerate indicates the frequency at which an obstacle policy with the given amount of rays can be loaded, executed, block-reduced and the resulting $(f,A)$ read back for execution. The red line indicates the maximum frequency ($20\, hz$) at which commonly available LiDARs (e.g. Ouster OS-1) can supply laser scans.}
  \label{fig:ray_cast_benchmark}
    \vspace{-1em}
\end{figure}
\Cref{fig:ray_cast_benchmark} shows the benchmarks for both policies. The map-based approach has static overhead caused by the map, which is most evident on the Jetson. The planner achieved frame rates vastly above the maximum sensor frequency on all tested architectures. The time needed for policy integration is nearly constant and, therefore, well predictable, mitigating the risk of bottlenecks during execution. 
\section{Conclusion}
This paper introduced a novel and practical method that enables using \acp{RMP} for obstacle avoidance on both voxelized maps and raw LiDAR observations. Massive parallel geometrical queries significantly improve planning performance. The resulting reactive planner performs as well as an optimization-based approach at a fraction of the computational cost. Computation acceleration is achieved by leveraging the parallelism of modern GPUs, which allows evaluating a staggering number of raycasting policies in a GPU-based voxelgrid map. Synthetic and real-world experiments demonstrate the qualities and performance of our approach. The method performs well in success-rate, smoothness, and compute utilization. Furthermore, we open source the implementation of the proposed method to enable future research on RMP-based navigation in challenging and cluttered environments.
%
%
%
\addtolength{\textheight}{-13.5cm}

\bibliography{bibliography} 
\bibliographystyle{ieeetr}

\end{document}